\documentclass{article}

\usepackage{arxiv}

\usepackage[utf8]{inputenc} 
\usepackage[T1]{fontenc}    
\usepackage{hyperref}       
\usepackage{url}            
\usepackage{booktabs}       
\usepackage{amsfonts}       
\usepackage{nicefrac}       
\usepackage{microtype}      
\usepackage{lipsum}		
\usepackage{graphicx}
\usepackage{natbib}
\usepackage{doi}
\usepackage{tabularx}
\usepackage{adjustbox}	
\usepackage{subcaption}
\usepackage{amsmath}
\usepackage{makecell}
\title{Innovative tokenisation of structured data for LLM training}

\author{ \href{https://orcid.org/0000-0002-3946-2048}{\includegraphics[scale=0.06]{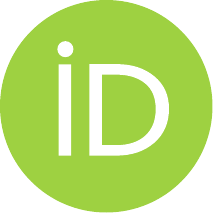}\hspace{1mm}Kayvan Karim}\\
	School of Mathematics and Computer Science\\
	Heriot-Watt University\\
	\texttt{k.karim@hw.ac.uk} \\
	\And
	\href{https://orcid.org/0000-0001-7393-0565}{\includegraphics[scale=0.06]{orcid.pdf}\hspace{1mm}Hani Ragab Hassen} \\
	School of Mathematics and Computer Science\\
	Heriot-Watt University\\
	\texttt{H.RagabHassen@hw.ac.uk} \\
    \And
	\href{https://orcid.org/0000-0003-0433-2152}{\includegraphics[scale=0.06]{orcid.pdf}\hspace{1mm}Hadj Batatia} \\
	School of Mathematics and Computer Science\\
	Heriot-Watt University\\
	\texttt{H.Batatia@hw.ac.uk} \\
}

\renewcommand{\shorttitle}{Innovative tokenisation of structured data for LLM training}

\hypersetup{
pdftitle={Innovative tokenisation of structured data for LLM training},
pdfsubject={Computer Science, Machine Learning},
pdfauthor={Kayvan Karim, Hani Ragab Hassen, Hadj Batatia},
pdfkeywords={Tokenization, Structured Data, LLMs, Data Representation,NIDS },
}

\begin{document}
\maketitle

\begin{abstract}
	Data representation remains a fundamental challenge in machine learning, particularly when adapting sequence-based architectures like Transformers and Large Language Models (LLMs) for structured tabular data. Existing methods often fail to cohesively encode the mix of numerical and categorical features or preserve the inherent structure of tables. This paper introduces a novel, hybrid tokenisation methodology designed to convert tabular data into a unified, sequential format suitable for LLM training. Our approach combines predefined fixed tokens to represent structural elements and low-cardinality categorical features, with a learned subword vocabulary using Byte-Pair Encoding (BPE) for high-cardinality and continuous values. We demonstrate the efficacy of this technique by applying it to a large-scale NetFlow dataset (CIDDS-001), preparing a corpus for a Network Intrusion Detection System (NIDS) foundation model. The evaluation shows that our method is highly efficient, processing over 31 million network flows in under five hours and achieving a significant data compression ratio of 6.18:1. This process resulted in a computationally manageable corpus of over one billion tokens, establishing a viable and generalisable pathway for training foundation models on structured data.
\end{abstract}


\section{Introduction}
Data representation is a foundational component of machine learning (ML) and deep learning (DL), as raw inputs must be transformed into numerical formats that models can process effectively. This chapter introduces a novel technique for tokenising structured data—specifically, tabular data such as Comma-Separated Values (CSV)— 
 designed for compatibility with transformer-based architectures.
Our proposed approach functions as a specialised tokeniser tailored to structured datasets. In the latter part of the chapter, we demonstrate the practical application of this tokeniser in the context of network log data, with a particular focus on NetFlow records. A CSV-based NetFlow dataset was tokenised using our method and subsequently employed to train a foundation model for Network Intrusion Detection Systems (NIDS).

The chapter is structured as follows: We begin with a comprehensive survey of existing data representation strategies, particularly those commonly used in natural language processing (NLP). This review highlights the limitations of current techniques when applied to structured data. We then present our tokenisation approach, detailing its design principles and advantages. The chapter concludes with an empirical evaluation and a discussion of the results obtained using our method.

\section{Learning Begins with Representation}
As Bengio et al. mentioned in their work, the effectiveness of ML algorithms, particularly in the domain of deep learning, is profoundly dependent on how data is represented \citep{yoshua_bengio_representation_2014}. Data representation, the process of transforming raw input data into a format suitable for ML models, is a cornerstone of the field. Different representations can expose or obscure the underlying factors of variation within the data, significantly impacting model performance \citep{yoshua_bengio_representation_2014}. Historically, considerable effort was dedicated to manual feature engineering, relying on domain expertise to craft informative representations. However, the advent of deep learning has shifted the paradigm towards representation learning, where models automatically discover complex patterns and hierarchical features directly from data \citep{zhong_overview_2016}.
This capability has driven breakthroughs across various domains, including computer vision, natural language processing (NLP), and time-series analysis \citep{noor_survey_2025}.

\section{Historical Overview of Data Representation Methods}
The journey of data representation in machine learning has evolved significantly over the past century, transitioning from manually crafted features and statistical methods to sophisticated, automatically learned representations \citep{zhong_overview_2016}. Understanding this history provides context for developing modern techniques, particularly those used in deep learning and large language models.

\subsection{Early Manual and Statistical Methods}

Before the rise of automated feature learning, representing data, especially text, relied heavily on statistical counts and manual encoding schemes. Most algorithms cannot directly process raw text or categorical data, necessitating conversion into numerical formats \citep{thakur_overview_2021}.

\textbf{One-Hot Encoding:} A fundamental technique for representing categorical data, where each category is mapped to a unique binary vector with a single $1$ and zeros elsewhere, for example, $[0,0,1,0]$. While simple, this method often leads to very high-dimensional and sparse vectors, especially with large vocabularies, a phenomenon known as the \textit{curse of dimensionality}. The lack of any notion of similarity between tokens (each token vector is orthogonal to others) \citep{siebers_survey_2022}.  For example, the words "cat" and "dog" would be two unrelated one-hot vectors, providing no hint that their meanings are similar.

\textbf{Bag-of-Words (BoW):} This model represents text documents as multisets of their words, disregarding grammar and word order but retaining word counts. A vocabulary is built from the entire corpus, and each document becomes a vector where dimensions correspond to vocabulary words and values represent their frequency. Like one-hot encoding, BoW produces sparse, high-dimensional vectors and loses sequential information \citep{asudani_impact_2023}.
For example, the encoding process of the two sentences below is:
\begin{verbatim}
s1 = "the cat sat on the mat"
s2 = "the dog sat on the log"
Vocabulary = ["the", "cat", "sat", "on", "mat", "dog", "log"]

\end{verbatim}

\begin{table}[h!]
\centering
\begin{tabular}{lccccccc}
\toprule
\textbf{Sentence} & \textbf{the} & \textbf{cat} & \textbf{sat} & \textbf{on} & \textbf{mat} & \textbf{dog} & \textbf{log} \\
\midrule
\textit{the cat sat on the mat} & 2 & 1 & 1 & 1 & 1 & 0 & 0 \\
\textit{the dog sat on the log} & 2 & 0 & 1 & 1 & 0 & 1 & 1 \\
\bottomrule
\end{tabular}
\caption{BoW encoding for two example sentences with repeated words.}
\end{table}

\textbf{Term Frequency-Inverse Document Frequency (TF-IDF):} TF-IDF is an improvement over simple BoW, TF-IDF weights terms based on their frequency within a document (TF) and their rarity across the entire corpus (IDF). This method assigns higher importance to words that are frequent in a specific document but infrequent overall, providing a better measure of a word's relevance to a particular document. Despite its effectiveness in information retrieval, TF-IDF still treats words as independent units and fails to capture semantic relationships \citep{asudani_impact_2023}.
\setlength{\jot}{4pt}
\[
\text{TF-IDF}(t, d) = \text{TF}(t, d) \times \log\left(\frac{N}{\text{DF}(t)}\right)
\]

\noindent\textbf{Where:}
\begin{align*}
t &:\ \text{term} \\
d &:\ \text{document} \\
N &:\ \text{total number of documents} \\
\text{DF}(t) &:\ \text{number of documents containing the term } t \\
\end{align*}

As an example, here is the TF-IDF for the following documents:
\begin{verbatim}
Doc 1: "sun rises in the east"
Doc 2: "moon rises in the night"
Doc 3: "stars twinkle"
Vocaulary = [sun, rises, in, the, east, moon, night, stars, twinkle]
\end{verbatim}

\begin{table}[h!]

\centering
\begin{tabular}{ll}
\toprule
\textbf{Term} & \textbf{Appears In} \\
\midrule
sun      & Doc 1 \\
rises    & Doc 1, Doc 2 \\
in       & Doc 1, Doc 2 \\
the      & Doc 1, Doc 2 \\
east     & Doc 1 \\
moon     & Doc 2 \\
night    & Doc 2 \\
stars    & Doc 3 \\
twinkle  & Doc 3 \\
\bottomrule
\end{tabular}
\caption{Step 1: Vocabulary extracted from the corpus}
\label{table:tf_idf_s1}
\end{table}

\begin{table}[h!]
\centering

\begin{tabular}{lccc}
\toprule
\textbf{Term} & \textbf{TF (Doc 1)} & \textbf{TF (Doc 2)} & \textbf{TF (Doc 3)} \\
\midrule
sun      & 0.20 & 0    & 0    \\
rises    & 0.20 & 0.20 & 0    \\
in       & 0.20 & 0.20 & 0    \\
the      & 0.20 & 0.20 & 0    \\
east     & 0.20 & 0    & 0    \\
moon     & 0    & 0.20 & 0    \\
night    & 0    & 0.20 & 0    \\
stars    & 0    & 0    & 0.50 \\
twinkle  & 0    & 0    & 0.50 \\
\bottomrule
\end{tabular}
\caption{Step 2: Term Frequency (TF) for each document}
\label{table:tf_idf_s2}
\end{table}

\begin{table}[h!]

\centering

\begin{tabular}{lc}
\toprule
\textbf{Term} & \textbf{Document Frequency (DF)} \\
\midrule
sun      & 1 \\
rises    & 2 \\
in       & 2 \\
the      & 2 \\
east     & 1 \\
moon     & 1 \\
night    & 1 \\
stars    & 1 \\
twinkle  & 1 \\
\bottomrule
\end{tabular}
\caption{Step 3: Number of documents each term appears in}
\label{table:tf_idf_s3}
\end{table}

\begin{table}[h!]
\centering
\begin{tabular}{lccc}
\toprule
\textbf{Term} & \textbf{TF-IDF (Doc 1)} & \textbf{TF-IDF (Doc 2)} & \textbf{TF-IDF (Doc 3)} \\
\midrule
sun      & 0.220 & 0.000 & 0.000 \\
rises    & 0.081 & 0.081 & 0.000 \\
in       & 0.081 & 0.081 & 0.000 \\
the      & 0.081 & 0.081 & 0.000 \\
east     & 0.220 & 0.000 & 0.000 \\
moon     & 0.000 & 0.220 & 0.000 \\
night    & 0.000 & 0.220 & 0.000 \\
stars    & 0.000 & 0.000 & 0.549 \\
twinkle  & 0.000 & 0.000 & 0.549 \\
\bottomrule
\end{tabular}
\caption{ Final TF-IDF scores for all terms across documents}
\label{table:tf_idf_s4}
\end{table}

Tables \ref{table:tf_idf_s1}, \ref{table:tf_idf_s2}, \ref{table:tf_idf_s3} and \ref{table:tf_idf_s4} demonstrate the process. 

\textbf{Co-Occurrence Vectors:} These vectors capture word relationships by counting how often words appear together within a defined context (e.g., a window of words). The resulting co-occurrence matrix represents words based on their shared contexts, offering a step towards capturing semantic similarity based on distributional properties \cite{thakur_overview_2021}.

These early methods laid the groundwork but were limited by their inability to capture deeper semantic nuances and reliance on high-dimensional, sparse representations.

\subsection{Early Feature Learning Algorithms}

Parallel to statistical text representations, early ML research explored algorithms designed to learn lower-dimensional, denser representations or features from data, aiming to capture the intrinsic structure without manual engineering \citep{zhong_overview_2016}.

\textbf{Linear Methods:} Techniques like Principal Component Analysis (PCA) focus on dimensionality reduction by finding orthogonal projections that maximise variance. Linear Discriminant Analysis (LDA) aimed to find projections that best separate predefined classes. Latent Semantic Analysis (LSA), using Singular Value Decomposition (SVD), sought to uncover latent semantic structures in text data by reducing the dimensionality of term-document matrices. Canonical Correlation Analysis (CCA) focuses on finding correlations between two sets of variables. These methods provided valuable data compression and analysis tools, but were limited to discovering linear relationships \citep{zhong_overview_2016}.

\textbf{Kernel Methods:} To capture non-linear structures, kernel methods like Kernel PCA (KPCA) and Generalised Discriminant Analysis (GDA) extend linear techniques using the "kernel trick," implicitly mapping data to higher-dimensional spaces where linear separation might be possible \citep{zhong_overview_2016}.

\textbf{Manifold Learning:} Algorithms such as Isometric Feature Mapping (Isomap), Locally Linear Embedding (LLE), and Locality Preserving Projections (LPP) emerged, assuming data lies on a lower-dimensional manifold within the high-dimensional observation space.3 These methods focused on preserving local neighbourhood structures or geodesic distances during dimensionality reduction, offering more powerful ways to model non-linear data structures \citep{zhong_overview_2016}.

These feature learning algorithms represented a significant step towards automated representation discovery, moving beyond simple statistical counts.

\subsection{The Transition to Neural Embeddings}
The limitations of traditional methods, particularly their inability to effectively capture semantic meaning and handle sparsity, motivated the development of neural network-based approaches. The key idea was to learn dense, low-dimensional vector representations (embeddings) where semantic relationships between words are reflected in the geometry of the embedding space \cite{asudani_impact_2023}.

\textbf{Word2Vec:} 
A significant breakthrough in natural language processing came with \textbf{distributed representations} in the form of learned embeddings, which map each token to a dense, low-dimensional vector so that semantically similar tokens lie close together rather than relying on sparse one-hot encodings.  Building on this paradigm, Mikolov et al.\cite{mikolov_distributed_nodate} introduced the Skip-gram model, a computationally efficient method that learns high-quality word embeddings by predicting surrounding context words; the paper also extends this approach to idiomatic phrases (e.g., “New York Times”) by treating them as single tokens, and shows that both word and phrase vectors exhibit additive compositionally \cite{mikolov_distributed_nodate}. For example,

\[
  \mathbf{vec}(\text{“Russia”}) + \mathbf{vec}(\text{“river”})
  \approx \mathbf{vec}(\text{“Volga River”}),
\]
 thereby capturing nuanced syntactic and semantic relationships.

\textbf{GloVe (Global Vectors for Word Representation):} Developed by Pennington et al. \cite{pennington_glove_2014}, GloVe combined the strengths of global matrix factorisation methods (like LSA) and local context window methods (like Word2Vec). It learns embeddings by factorising a global word-word co-occurrence statistics matrix, proving efficient and effective \cite{pennington_glove_2014}.

\textbf{fastText}: After Mikolov moved to Facebook AI Research, they developed fastText as an extension of Word2Vec. This method incorporates subword information by representing words as bags of character n-grams. This allows it to generate embeddings for out-of-vocabulary (OOV) words and better capture morphological information, making it particularly useful for morphologically rich languages \cite{bojanowski_enriching_2017}. Additionally, fastText was released by Facebook as an open-source library \cite{fastText}. \\

This shift towards learned, dense embeddings represented a fundamental change in NLP, moving away from feature engineering towards representation learning powered by neural networks \cite{naseem_comprehensive_2020}. These embeddings provided the foundation upon which more complex models, like Transformers and LLMs, would later build.

\section{Tokenisation for Transformers and Large Language Models}
Modern NLP is dominated by Transformer-based architectures and Large Language Models (LLMs) \citep{minaee_large_2025}. A crucial preliminary step for these models is tokenisation: the process of converting raw input text into a sequence of numerical IDs that the model can ingest \citep{schmidt_tokenization_2024}. Unlike early methods that often operated on whole words, Transformers and LLMs almost universally rely on subword tokenisation algorithms.

\subsection{The Need for Subword Tokenisation}
Operating purely at the word level presents several challenges for large-scale language models. The following is a list of some of the most important challenges. 

\textbf{Out-of-Vocabulary (OOV) Words:} Any fixed vocabulary will inevitably encounter words not seen during training (e.g. rare words, new names, typos). Word-level models struggle to handle these OOV terms, often resorting to a generic \verb#<UNK># (unknown) token, which leads to information loss \citep{rahman_towards_2024}.

\textbf{Morphology:} Languages with rich morphology (e.g. Arabic, Turkish, German) generate many word forms from a single root. Representing each form as a unique token is inefficient and fails to capture the underlying relationships between related words (e.g., "run", "running", "ran") \citep{alyafeai_evaluating_2021}.

\textbf{Vocabulary Size:} A purely word-based vocabulary for the vast amounts of text used to train LLMs would become excessively large, increasing model size and computational cost \citep{qarah_comprehensive_2024}.

Subword tokenisation addresses these issues by breaking words into smaller, meaningful units (subwords). Common words might remain intact, while rare words are segmented into known subwords (e.g., "tokenisation" $->$ "token", "ization"). This method allows models to handle OOV words, capture morphological patterns, and maintain a manageable fixed-size vocabulary \citep{rahman_towards_2024}.

\subsection{Common Subword Tokenisation Algorithms}
Several algorithms have become standard for subword tokenisation, primarily from data compression techniques \citep{schmidt_tokenization_2024}.

\label{BPE_Exp}\textbf{Byte-Pair Encoding (BPE):} BPE was first introduced as a data compression method in 1994 by Philip Gage \citep{gage_new_1994}. It was then popularised in NLP and used for neural machine translation by Sennrich et al. \citep{sennrich_neural_2016}. BPE operates bottom-up, initialising the vocabulary with individual characters (or bytes) and then iteratively counting all adjacent pairs of symbols in the training corpus and merging the most frequent pair into a single new symbol, adding it to the vocabulary. This merge process repeats until a predefined vocabulary size is reached. For encoding (segmentation method), BPE applies the learnt merge rules greedily to the new text \citep{schmidt_tokenization_2024}. BPE handles OOV words by breaking them into known subwords/characters. It is language-agnostic and can capture some morphological variations due to its character-level operation. BPE is used in models like the GPT series \citep{rahman_towards_2024}. 

\label{WordPiece}\textbf{WordPiece:} Similar to BPE, WordPiece builds a vocabulary iteratively, starting from individual characters \citep{schmidt_tokenization_2024}. However, WordPiece merges the pair, which maximises the likelihood of the training data given the current vocabulary, instead of merging the most frequent pair. This likelihood-based criterion distinguishes it from BPE's frequency-based approach. In addition, it is language-agnostic and reduces vocabulary size. Encoding is typically greedy, finding the longest known subword prefix in the remaining text. It often uses a prefix (e.g., \verb|##|) to denote subwords that are part of a larger word \citep{qarah_comprehensive_2024}. WordPiece is used in models such as BERT and DistilBERT \citep{rahman_towards_2024}.

\textbf{Unigram Language Model:} This subword segmentation algorithm, proposed by Kudo \citep{kudo_subword_2018}, is also presented for neural machine translation. Unigram takes a top-down approach. It starts with a large initial vocabulary of candidate subwords (e.g., all substrings or frequent character n-grams) and assumes each subword occurs independently (unigram assumption). It then iteratively removes tokens from the vocabulary, prioritising the removal of those whose elimination causes the smallest decrease in the overall likelihood of the training corpus, calculated using the unigram model probabilities. This pruning continues until the target vocabulary size is met \citep{schmidt_tokenization_2024}. Encoding is probabilistic; given a text, it typically uses the Viterbi algorithm \cite{viterbi_error_1967} to find the most likely sequence of subword tokens based on their learned probabilities. This allows for multiple possible segmentations of the same word, which can be used for regularisation during training (subword sampling) \citep{schmidt_tokenization_2024}.

\textbf{SentencePiece} SentencePiece is not a distinct tokenisation algorithm itself but rather a software library and framework that implements both BPE and Unigram algorithms \citep{sentencepiece}. A key feature is its language-independent design. It treats input text as a raw sequence of Unicode characters. It trains the tokeniser directly from these raw sentences, eliminating the need for language-specific pre-tokenisation (like splitting on whitespace). It explicitly handles whitespace, often by converting it into a special meta-symbol (e.g., "\_" (U+2581)), which simplifies detokenisation and improves consistency. SentencePiece also supports subword regularisation techniques, such as BPE dropout. It is used in models such as XLNet and ALBERT \citep{rahman_towards_2024}.

\subsection{Comparative Analysis and Challenges}
The choice of tokenisation algorithm is increasingly recognised as a critical modelling decision rather than a simple preprocessing step. Research shows significant variability in how different tokenisers segment the same text, which directly impacts downstream model performance, sequence length, and the learned representations \citep{schmidt_tokenization_2024}.

These algorithms differ fundamentally in their construction approach (bottom-up BPE/WordPiece vs. top-down Unigram), the methods for merging or pruning tokens (frequency vs. likelihood), and the segmentation strategy (greedy vs. probabilistic) \citep{schmidt_tokenization_2024}. BPE prioritises compression based on frequency, while WordPiece and Unigram incorporate statistical likelihood, potentially yielding more linguistically plausible subwords, though BPE can sometimes create non-meaningful units \citep{alyafeai_evaluating_2021}.
 Unigram's probabilistic nature allows for exploring multiple segmentations, offering a form of regularisation. SentencePiece provides an end-to-end, language-agnostic system that handles normalisation and raw text input directly \citep{rahman_towards_2024}.

 The initial hypothesis that simply minimising the number of tokens leads to better performance is being questioned. Studies show that factors such as pre-tokenisation rules (e.g., how whitespace or punctuation is handled initially) and the specific vocabulary construction process also play significant roles \citep{schmidt_tokenization_2024}.
 No single tokenisation technique consistently outperforms others across all scenarios; the optimal choice depends heavily on the specific task, the size and nature of the dataset, and the morphological complexity of the language(s) involved \citep{alyafeai_evaluating_2021}.  Evaluating tokenisers requires considering their impact on downstream tasks, not just intrinsic properties like compression rate \citep{schmidt_tokenization_2024}.

Despite advancements, tokenisation remains a potential point of failure, even for state-of-the-art LLMs like GPT-4. When an LLM's internal tokenisation of an input sentence significantly misaligns with human understanding or the intended meaning, it can lead to inaccurate, illogical, or nonsensical responses. This suggests that the tokenisation layer can act as a bottleneck, preventing the model from correctly interpreting certain inputs, regardless of its internal reasoning capabilities. Researchers are developing adversarial datasets (e.g., Adversarial Dataset for Tokeniser) specifically designed to probe and expose these tokenisation vulnerabilities  \citep{wang_tokenization_2025}.

Tokenisation is particularly crucial for multilingual LLMs (MLLMs) \citep{qin_survey_2025}. Frameworks like SentencePiece, designed to be language-agnostic and handle raw Unicode, are often preferred. However, achieving optimal segmentation across multiple languages, especially in cross-lingual settings, remains challenging, and standard subword methods might still produce suboptimal results \citep{alyafeai_evaluating_2021}.

Active research focuses on developing improved tokenisation algorithms (e.g., PathPiece aims for minimal token segmentation), gaining a deeper understanding of the complex interplay between tokenisation choices and model architecture/performance, and creating more robust tokenisers that are less susceptible to misinterpretations and adversarial inputs.
The tension between optimising for data compression versus linguistic meaningfulness continues to be a key area of exploration \citep{schmidt_tokenization_2024}. 

The table \ref{tab:tokeniser_compare} summarises the key characteristics of the discussed subword tokenisation algorithms:
\begin{table}[h]
    \scriptsize 
    \centering
    \begin{adjustbox}{width=\textwidth}
    \begin{tabularx}{\textwidth}{lXXXX}
    \toprule
         \textbf{Feature} & \textbf{BPE (Byte-Pair Encoding)} & \textbf{WordPiece} & \textbf{Unigram LM} & \textbf{SentencePiece (Framework)} \\
    \midrule
         Core Mechanism & Bottom-up merging & Bottom-up merging & Top-down pruning & Implements BPE \& Unigram \\
         Merge/Prune Rule & Most frequent adjacent pair & Pair maximising data likelihood & Pair minimising likelihood loss & Depends on chosen algorithm (BPE/Unigram) \\
         Segmentation & Greedy application of merge rules & Greedy longest prefix matching & Probabilistic (e.g., Viterbi) & Depends on chosen algorithm (BPE/Unigram) \\
         Handling Raw Text & Typically requires pre-tokenisation & Typically requires pre-tokenisation & Can work from substrings & Handles raw Unicode directly, no pre-tokenisation needed\\
         Key Strengths & Simple, effective compression & Likelihood-based, used in BERT & Probabilistic, allows sampling & Language-agnostic, end-to-end, handles whitespace \\
         Key Weaknesses & Can create non-meaningful units, greedy & Greedy segmentation & Can be complex to implement & Framework overhead, choices still matter \\
         Common Models & GPT series & BERT, DistilBERT & Used in SentencePiece & XLNet, ALBERT, T5, many MLLMs \\
    \bottomrule
    \end{tabularx}
    \end{adjustbox}
    \caption{Comparison of Subword Tokenisation Algorithms}
    \label{tab:tokeniser_compare}
\end{table}

\section{Representation for Structured Data  (CSV, Tables) in ML and LLMs}
While LLMs have revolutionised NLP, applying them effectively to structured data, such as the tables found in CSV files or databases, presents unique challenges and necessitates specialised representation techniques  \citep{fang_large_2024}. Tabular data is ubiquitous across various domains, including finance, healthcare, and scientific research, making its effective representation crucial \citep{ren_deep_2025}.

Representing tabular data effectively poses several inherent difficulties for both traditional ML and modern deep learning models, including LLMs. Some of these challenges include heterogeneity, sparsity and data quality, complex dependencies, structural mismatches with sequence models, and scalability. We will explain these challenges below.

\textbf{Heterogeneity:} Tables typically contain diverse data types within their columns – numerical (continuous or discrete), categorical (nominal or ordinal), binary, and sometimes free text \citep{ren_deep_2025}.
Numerical features often exhibit complex, non-Gaussian distributions (skewness, heavy tails, multimodality), while categorical features can have very high cardinality (many unique values) or suffer from severe imbalances. Each type requires appropriate encoding and handling \citep{fang_large_2024}.

\textbf{Sparsity and Data Quality}: Real-world tabular datasets frequently suffer from missing values due to various reasons (privacy, measurement errors, optional fields). Furthermore, class imbalance is common, where certain outcomes or categories are vastly underrepresented. These issues lead to sparse data matrices and long-tailed distributions, complicating model training and potentially biasing outcomes \citep{shi_comprehensive_2025}.

\textbf{Complex Dependencies:} Meaningful insights often lie in the intricate, non-linear interactions between different columns (features)  \citep{ren_deep_2025}. Capturing these high-order dependencies is essential for accurate modelling but challenging for simpler models like linear regression or decision trees \citep{jiang_representation_2025}.

\textbf{Structural Mismatch with Sequence Models:} LLMs are fundamentally sequence processors pre-trained on vast amounts of linear text  \citep{lu_large_2025}. Tables, however, possess an inherent two-dimensional structure with rows representing instances and columns representing attributes. Directly applying LLMs by simply linearising a table can obscure this structure, making it difficult for the model to understand row-column relationships, schema information (column names and types), and the context-dependent meaning of cell values (e.g., the value '10' means different things in an 'Age' column versus a 'Quantity' column) \citep{shi_comprehensive_2025}.

\textbf{Scalability:} Tables can range from small spreadsheets to massive database extracts with millions of rows or hundreds of columns \citep{balaka_pneuma_2025}. This poses challenges for LLMs with fixed context window limits, which may not be able to process an entire large table at once \citep{koloski_llm_2025}. It also creates scalability issues for storage and indexing, particularly in retrieval-augmented generation (RAG) systems where table representations need to be efficiently stored and searched \citep{balaka_pneuma_2025}.

\subsection{Techniques for Tabular Data (Pre-LLM / Specialised DL)}

Before the widespread application of LLMs to tabular data, significant research focused on developing specialised deep learning architectures and techniques tailored to the unique characteristics of tabular data \citep{ren_deep_2025}.

 DNNs offered several potential advantages over classical ML methods (e.g., gradient-boosted trees, random forests) for tabular data. These include the inherent ability to model complex, high-order, non-linear feature interactions through hierarchical representations, the capacity for end-to-end learning directly from raw or minimally processed features, reducing reliance on manual feature engineering, flexibility in integrating tabular data with other modalities (images, text, time-series) within multi-modal pipelines, adaptability to dynamic environments and online learning scenarios via gradient-based optimisation, and the potential for long-term knowledge transfer across related tasks or datasets \citep{jiang_representation_2025}.

 A crucial first step in applying DNNs to tabular data is embedding the heterogeneous input features into a common high-dimensional vector space  \citep{koloski_llm_2025}. 
Categorical features are typically handled using learnable embedding layers, similar to word embeddings in NLP. Numerical features require different strategies, such as binning, applying transformations (like normalisation or scaling), or using specialised numerical encoding layers that project them into the embedding space.

Recognising that standard architectures like Multi-Layer Perceptrons might not optimally capture tabular structures, researchers developed specialised DNNs. Examples include Transformer-based models adapted for tabular data, such as FT-Transformer (Feature Tokeniser Transformer), which utilises attention mechanisms to weigh the importance of different features for each sample. ResNet-like architectures have also been adapted \citep{koloski_llm_2025}.

Additionally, when relationships between data points (rows) can be modelled as a graph (either explicitly provided or inferred from features), Graph Neural Networks (GNNs) have been employed for tasks like node classification or link prediction on tabular data represented as graphs \citep{li_curriculum_2024}. 
Several surveys focus specifically on these specialised deep-learning approaches for tabular data \citep{ren_deep_2025}.

\subsection{Bridging Tabular Data and Large Language Models}

The remarkable success of LLMs in NLP has developed interest in adapting them for tabular data tasks, aiming to leverage their vast pre-trained knowledge and emergent reasoning capabilities \citep{fang_large_2024}.

LLMs offer potential solutions to some long-standing challenges in tabular modelling. Their pre-training on diverse text corpora equips them with world knowledge and semantic understanding that could be beneficial for interpreting feature names and values  \citep{koloski_llm_2025}.
They naturally handle textual features within tables. Transforming tabular data into natural language sequences can mitigate the "curse of dimensionality" associated with high-cardinality categorical features often encoded using one-hot methods. Furthermore, the emergent abilities of LLMs, such as few-shot learning and step-by-step reasoning (Chain-of-Thought), open possibilities for tackling complex tabular reasoning, question answering, and data generation tasks with less task-specific training data \citep{fang_large_2024}.

A primary challenge is converting the 2D table structure into a 1D sequence suitable for LLM input. This process, known as serialisation or linearisation, involves representing the table's content and structure as a sequence of text. Several methods have been explored:

\textbf{Simple Delimiter-Separated:} Using standard formats like CSV (Comma-Separated Values) or TSV (Tab-Separated Values), often including the header row \citep{fang_large_2024}.

\textbf{Markdown:} Employing Markdown table syntax with pipes (\textbar ) separating columns and hyphens (-) separating the header \citep{fang_large_2024}.

\textbf{JSON Formats:} Representing the table as a JSON object. This can be column-based (keys are column names, values are lists of cell contents) or row-based (a list of JSON objects, each representing a row with key-value pairs for columns) \citep{fang_large_2024}.

\textbf{HTML}: Using standard HTML \verb|<table>, <tr>, <the>, <td>| tags to encode the structure \citep{fang_large_2024}.

\textbf{Key-Value Pairs:} Concatenating column names and cell values for each cell or row (e.g., \verb|col1:val1;col2:val2...|) \citep{fang_large_2024}.

\textbf{Natural Language Sentences:} Using templates to convert each row into a descriptive sentence (e.g., "For row 1, the Name is Helen and the Age is 47") \citep{fang_large_2024}.

\textbf{Bracketed Representations:} Using brackets (\verb|``|  or \verb|{}|) to delineate rows or cells within the linearised sequence explicitly \cite{deng_tables_2024}.

\textbf{Code Representations:} Representing the table as code that would generate it, e.g., Python code using Pandas (\verb|pd.DataFrame({...})| ) \citep{fang_large_2024}.

Research shows that the choice of serialisation significantly impacts LLM performance \citep{deng_tables_2024}. Simple linearisation might lose structural information, while formats like JSON or Markdown attempt to preserve it better \citep{fang_large_2024}. Bracketed representations have shown promise, potentially because LLMs encounter similar structures in code during pre-training. More capable LLMs, such as GPT-4, appear relatively robust to different text formats, while others benefit more from structure-preserving representations. \citep{deng_tables_2024}. There is a trade-off between preserving structure, conciseness, and LLM familiarity with the format. A comparison of these methods is presented in  \ref{tab:Serialisation_Method}. 

\begin{table}[h]
    \scriptsize  
    \centering
    \begin{adjustbox}{width=\textwidth}
    \begin{tabularx}{\textwidth}{XXXXXX}
    \toprule
         \textbf{Method} & \textbf{Description} & \textbf{Example} & \textbf{Key Characteristic} & \textbf{Potential Pros} &  \textbf{Potential Cons} \\
     \midrule
        X-Separated (CSV) & Columns separated by delimiter (e.g., comma), rows by newline & \verb|name,age,47| & Simple, common & Concise, widely used & Minimal structure encoding, ambiguity with delimiter \\
 \midrule
        Markdown & Uses Markdown table syntax. & 
        \makecell[l]{\texttt{| name  | age |} \\ \texttt{|------|-----|} \\ \texttt{| helen | 47 |}} 
        & Human-readable structure & Preserves visual layout, some structure & Can be verbose \\
 \midrule
        JSON (Row-based) & List of JSON objects, each object is a row with key-value pairs. & 
        \texttt{[{ "name": "helen", "age": 47 }]}
        & Explicit key-value structure & Good structure preservation (row-level) & Verbose, potentially long sequences \\
 \midrule
        JSON (Column-based) & JSON object where keys are columns, values are lists of cells. & 
        \texttt{{ "name": ["helen"], "age": [47] }} 
        & Explicit column structure & Groups column data together & Less intuitive row access, potentially long lists \\
 \midrule
        HTML & Uses HTML \texttt{<table>}, \texttt{<tr>}, \texttt{<td>} tags. & 
        \texttt{<table><tr>
        <td>helen</td>
        <td>47</td></tr>
        </table>}  
        & Standard web format & Explicit structure, common in web data & Very verbose, many non-data tokens \\
 \midrule
        Attribute-Value Pairs & Concatenated ``column: value'' pairs per row. & 
        \texttt{name:helen ; age:47}
        & Key-value pairing & Explicit column context for each value & Can lose overall table structure, repetitive \\
 \midrule
        Sentences & Rows converted to natural language using templates. & 
        \texttt{name is helen, age is 47} 
        & Natural language format & Highly readable, leverages NLP capabilities & Template design crucial, potential ambiguity \\
 \midrule
        Bracketed & Uses brackets \texttt{[]} or \texttt{\{\}} to delineate rows/cells. & 
        \texttt{[[name, age], [helen, 47]]}
        & Explicit grouping & Can improve structure perception for LLMs & Syntax might vary, potentially less standard \\
 \midrule
        DFLoader (Code) & Python code (e.g., Pandas) to create the dataframe. & 
         \texttt{pd.DataFrame
        (\{'name': ['helen'], 
        'age': [47]\})}
        & Code representation & Leverages code understanding of LLMs & Requires code execution context, specific \\

    \bottomrule
    \end{tabularx}
    \end{adjustbox}
    \caption{Comparison of Table Serialisation Methods for LLMs}
    \label{tab:Serialisation_Method}
\end{table}

\subsection{LLM Embeddings for Tabular Features}

An alternative or complementary approach is to use LLMs as powerful feature extractors, injecting their semantic understanding into other ML models. Instead of asking the LLM to perform the final prediction or task, this method utilises the LLM to generate rich embeddings for the tabular data, which are then fed as input features into a downstream classifier or regressor (which could be a traditional model, such as XGBoost, or a specialised deep tabular model). This typically involves converting rows or feature-value pairs into textual descriptions (e.g., "Feature 'Age' has value 47"). These text snippets are fed into a pre-trained LLM (like RoBERTa, GPT-2, BGE, Llama3) to obtain contextual embeddings. These embeddings, potentially after projection or adaptation via a small MLP, serve as enriched features for the downstream task model.
This "context injection" leverages the LLM's pre-trained knowledge about language and potentially the world, enhancing the representation of tabular features, especially categorical ones or those with descriptive names \citep{koloski_llm_2025}.

Other studies have shown that this approach can improve performance over baseline models (MLP, ResNet, FT-Transformer) on specific datasets, particularly those with class imbalance or limited features. Feature importance analyses often reveal that LLM-derived embeddings rank highly \citep{kasneci_enriching_2024}.

However, the effectiveness depends on the chosen LLM, the nature of the dataset (purely numerical data might benefit less), and whether the data domain is well-represented in the LLM's training data \citep{koloski_llm_2025}. Using larger LLMs does not always guarantee better performance as feature extractors.

\subsection{Specialised Tabular Embeddings and Models for LLMs }

Limitations of direct application or simple embedding extraction have led research to explore more specialised ways to integrate LLMs with tabular data. Here we list the most common ones:

\textbf{Fine-tuning Embedders for Tabular RAG}: For retrieval-augmented generation involving tables, standard text embedding models may struggle. Approaches like the Tabular Embedding Model (TEM) involve fine-tuning embedding models specifically on tabular data structures and queries to improve retrieval accuracy for table-based RAG applications \citep{khanna_tabular_2024}.

\textbf{LLM-based Encoders:} Novel tabular encoder architectures are being designed that incorporate principles from LLMs or address specific challenges. For instance, methods employ multi-tier partitioning based on power-law dynamics to handle sparse, high-cardinality categorical features, adaptive quantisation techniques to impose priors on numerical continuity for better numerical reasoning, and distinct processing pathways for core data columns versus meta-information columns \citep{raman_scalable_2024}.

\textbf{Table Representation for Retrieval (RAG)}: Systems like \textbf{Pneuma} focus on creating efficient and effective table representations specifically for discovery and retrieval tasks. Pneuma uses LLMs to generate natural language narrations of the table schema (providing meaningful descriptions even for cryptic column names) and summaries of sampled rows, preserving key information while remaining scalable compared to embedding entire tables or splitting them into excessive row-level snippets. This representation is then encoded using standard embedding techniques for efficient vector search, combined with full-text search and LLM-based re-ranking \citep{balaka_pneuma_2025}. 

\textbf{Dedicated Table LLMs:} A significant trend is the development of LLMs specifically fine-tuned or instruction-tuned on large, diverse datasets of table-related tasks \citep{lu_large_2025}. Models, such as TableLLM\citep{zhang_tablellm_2025} and TableLlama \citep{zhang_tablellama_2024}, are trained to handle tasks like table question answering, data manipulation (including query, update, merge, and chart generation), and understanding tables embedded within documents (Word/PDF) or spreadsheets (Excel/CSV). This specialisation aims to equip the LLMs with a deeper intrinsic understanding of table structures and operations.

\subsection{Multimodal Approaches (Image-based Table Representation)}
 With the rise of Multimodal LLMs (MLLMs) capable of processing images, representing tables visually has become an alternative strategy \citep{lu_large_2025}. 

In this method, instead of serialising the table to text, an image of the table is provided as input to an MLLM, such as GPT-4V or Gemini. Variations include using a plain table image (Vanilla-V), colouring columns differently (Column-Colour), or colouring rows differently (Row-Colour) to provide visual cues \citep{deng_tables_2024}.

MLLMs demonstrate a capacity to reason over table images, sometimes achieving performance comparable or even superior to text-based representations, particularly on tasks involving complex numerical relationships or where visual layout is important \citep{deng_tables_2024}.

Prompting techniques, such as the Chain-of-Thought method, are beneficial for image-based reasoning. Visual cues, such as colour, can influence performance, with row distinctions sometimes proving more helpful than column distinctions; however, the effect varies between models. Performance might also depend on whether the MLLM encountered similar tables predominantly as text or images during its pre-training \citep{deng_tables_2024}.

\section{Future Outlook}

The field is moving towards more nuanced and hybrid strategies. Purely specialised deep tabular models face generalisation challenges, while applying generic LLMs directly often fails to capture the specific structural properties of tables. The most promising avenues involve intelligently combining the strengths of both worlds: using LLMs to enrich tabular models with semantic context, fine-tuning LLMs with table-specific knowledge, or designing novel architectures inspired by both paradigms. Success in complex tabular tasks, such as reasoning, manipulation, and generation, appears heavily contingent on developing representations that effectively bridge the structural gap between tables and the sequential processing nature of LLMs. This suggests a shift from seeking a single, universal representation to developing more task-aware representations tailored to specific downstream applications, such as retrieval, prediction, or generation.

\section{A Novel Tokenisation Approach for Tabular Data}
Effectively representing tabular data is a primary challenge that limits the performance of advanced machine learning models, particularly sequence-based architectures such as Transformers. Existing methods often struggle to cohesively encode the mix of numerical and categorical features inherent in structured data. To address this gap, we introduce a novel tokenisation methodology that converts tabular entries into a unified, sequential format. This approach, while broadly applicable to any tabular dataset, is demonstrated explicitly in this work on NetFlow data to showcase its efficacy in a complex, real-world scenario.

\section{Methodology}
To train our tokeniser and prepare the data for a sequence-based model, we first preprocess the raw NetFlow data. Then, the preprocessed data is used to construct a hybrid token vocabulary. Finally, the new tokenisation is used to tokenise the entire dataset into a single numerical sequence.

\subsection{Data Preprocessing and Feature Engineering}
The primary goal of the preprocessing stage is to transform the raw, multi-file dataset into a single, chronologically ordered, and feature-rich DataFrame suitable for tokenisation. This involves three key steps: \textbf{data aggregation}, \textbf{temporal ordering} with feature engineering, and final \textbf{data type correction}.

First, data aggregation is performed to consolidate the dataset. The CIDDS-001 dataset \citep{ring_technical_2017} is distributed across four separate CSV files, each containing one week of internal network traffic. These files are systematically read and loaded into memory using the Pandas library. They are then concatenated into a single, unified DataFrame. This step creates a comprehensive dataset of \textbf{31,287,933} individual network flows, representing one month of continuous network activity.

Second, \textbf{temporal ordering and feature engineering} are performed to imbue the dataset with a sequential context, which is critical for sequence-aware models. The \verb|`Date first seen`|  column, initially represented as a string, is converted into a proper datetime object. This allows for accurate chronological sorting of the entire DataFrame, ensuring that the network flows are arranged in the exact order they were observed. This temporal sequence is fundamental for the model to learn time-based patterns, such as the progression of an attack.

With the data sorted, a new and vital feature, \textbf{DeltaTime}, is engineered. This feature is calculated as the time difference, in total seconds, between each flow and the one immediately preceding it \verb|(.diff().dt.total_seconds())|. The \verb|DeltaTime| feature explicitly encodes the inter-arrival time between consecutive network events, a potentially powerful indicator for detecting anomalies like network scans or denial-of-service attacks, which often manifest as rapid bursts of traffic with very small delta times. As a result of this operation, the very first flow in the dataset has no preceding flow, resulting in a NaN (Not a Number) value for its DeltaTime. This is an expected outcome and is handled in the subsequent tokenisation step by converting it to the literal string 'nan'. Following the creation of this feature, the original Date first seen column is dropped from the DataFrame. This is a deliberate choice to prevent data redundancy, as the temporal information is now implicitly captured by the row order and explicitly by the DeltaTime feature.

\begin{figure}
    \centering
    \includegraphics[width=1\linewidth]{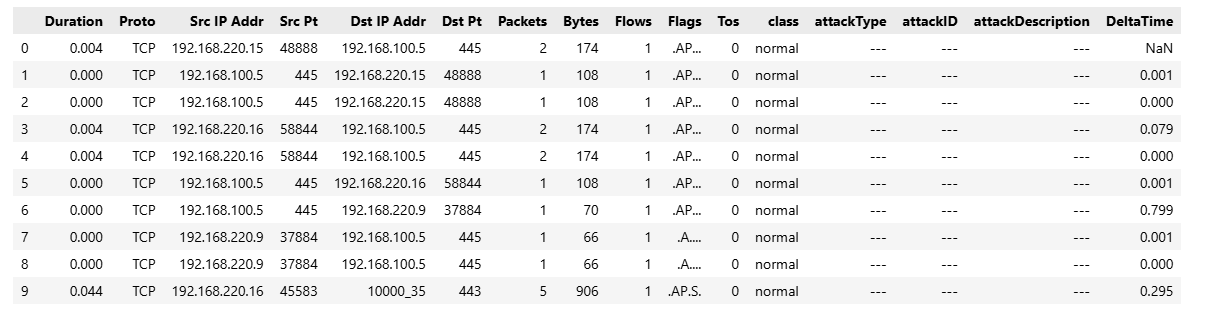}
    \caption{A Sample of Preprocessed and Feature-Engineered NetFlow Data}
    \label{fig:netflow_sample}
\end{figure}

Following the creation of this feature, the original Date first seen column is dropped from the DataFrame. This is a deliberate choice to prevent data redundancy, as the temporal information is now implicitly captured by the row order and explicitly by the DeltaTime feature. A sample of the final output can be seen in Figure \ref{fig:netflow_sample}.

Finally, a data type correction step is performed as a sanity check to ensure data integrity. Columns that represent numerical values but may have been loaded as objects or strings, such as \verb|Dst Pt| (Destination Port), are explicitly cast to the \verb|integer| data type. This prevents potential errors in later processing and ensures all data conforms to the expected schema before tokenisation.

\label{sec:vocab}
\subsection{Hybrid Vocabulary Construction}
At the heart of our approach is a custom-trained tokeniser, \verb|NIDSTokenizer|, which employs a hybrid vocabulary strategy to represent the diverse data types in NetFlow records efficiently. The vocabulary is composed of two distinct types of tokens: fixed tokens and learned subword tokens.

\textbf{Fixed Tokens:} These are predefined tokens that represent \textbf{structural elements} of the data or low-cardinality \textbf{categorical features}.
\begin{itemize}
    \item \textbf{Structural Tokens:} Special tokens like \texttt{<|SRCIP|>}, \texttt{<|DSTPORT|>}, and \texttt{<|ROW|>} are used to denote the start of a specific feature or the end of a data row. These tokens allow the model to understand the inherent structure of the tabular data within a linear sequence.
    \item \textbf{Categorical Tokens:} For features with a small, fixed set of possible values (e.g., \verb|Proto|, \verb|Flags|, \verb|class| ), each unique value is mapped to a distinct token (e.g., \texttt{<TCP>}, \texttt{<UDP>}). This includes placeholder or null-equivalent values, such as \texttt{<--->} for the \verb|attackType| `feature, ensuring they are treated as unique and consistent categorical inputs. This ensures a consistent and compact representation for these features.
\end{itemize}

\textbf{Learned Subword Tokens}: For features with high cardinality or continuous values (e.g., IP addresses, port numbers, duration, bytes), a simple fixed vocabulary would be impractically large. To handle this, we employ \textbf{Byte-Pair Encoding (BPE)} as explained in \ref{BPE_Exp} for the subword tokenisation algorithm. The BPE merge rules are learned iteratively from the corpus of unique values. The training process is conducted sequentially on distinct feature types—first on IP addresses, then on port numbers, and finally on the remaining numerical and continuous features—to allow the tokeniser to learn subword patterns relevant to each data category. For example, instead of having a unique token for every IP address, the BPE model learns to represent common numerical patterns and character sequences (like \verb|192.168.|) as single tokens, breaking down rare or unseen IPs into smaller, known subword units. This approach maintains a fixed, manageable vocabulary size while retaining the ability to represent a virtually infinite set of values. This approach maintains a fixed, manageable vocabulary size while retaining the ability to represent a virtually infinite set of values.

The tokeniser is trained by first initialising the vocabulary with all fixed tokens. Subsequently, the BPE training is performed iteratively on the unique string values from the high-cardinality columns, learning a specified number of merge rules (e.g., 1000 merges per column type) to build the subword vocabulary. For our data, we used 1000 merges per column.

\begin{figure}[htbp]
  \centering

  \begin{minipage}[b]{0.3\textwidth}
    \includegraphics[width=\textwidth]{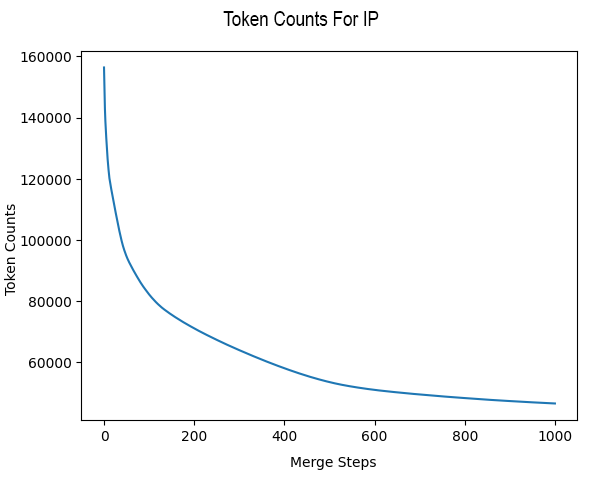}
  \end{minipage}
  \hfill
  \begin{minipage}[b]{0.3\textwidth}
    \includegraphics[width=\textwidth]{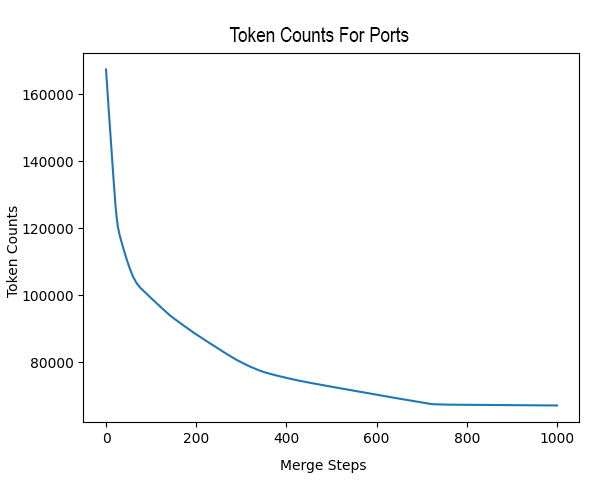}
  \end{minipage}
  \hfill
  \begin{minipage}[b]{0.3\textwidth}
    \includegraphics[width=\textwidth]{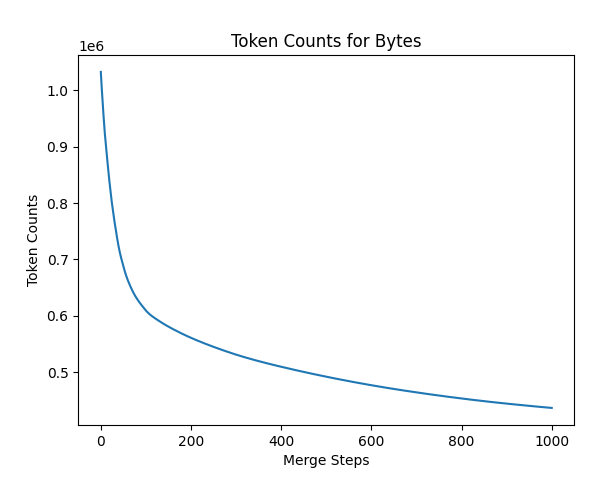}
  \end{minipage}

  \vspace{1em}

  \begin{minipage}[b]{0.3\textwidth}
    \includegraphics[width=\textwidth]{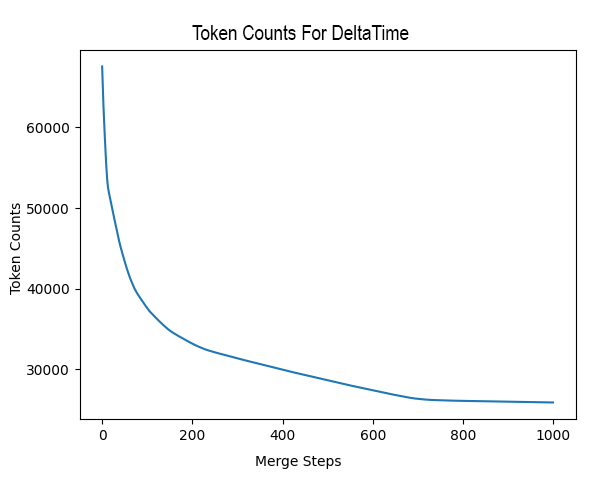}
  \end{minipage}
  \hfill
  \begin{minipage}[b]{0.3\textwidth}
    \includegraphics[width=\textwidth]{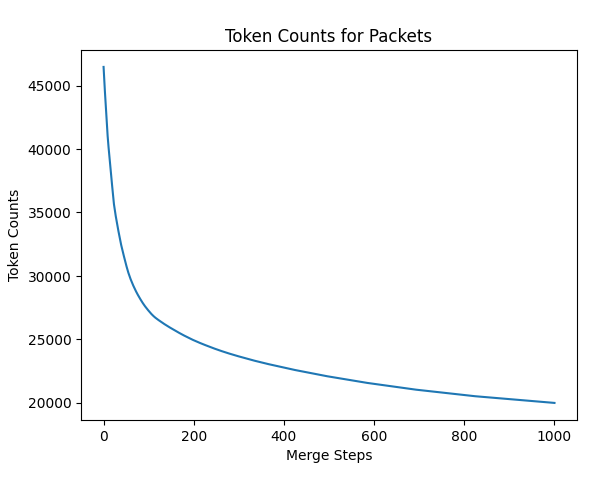}
  \end{minipage}
  \hfill
  \begin{minipage}[b]{0.3\textwidth}
    \includegraphics[width=\textwidth]{Chapters/Chapter4/imgs/TC_DetaTimeE.png}
  \end{minipage}

  \caption{Token counts per mergs steps for BPE columns}
  \label{fig:tokencounts}
\end{figure}

As shown in Figure \ref{fig:tokencounts}, the token counts plateau after 1,000 merges. After all the merges, the total number of vocabulary reached 4,241.

\subsection{Final Tokenisation and Corpus Generation}

With the trained tokeniser, the preprocessed  DataFrame is converted into a text corpus. As described previously, each row is transformed into a single string by concatenating the structural token for each column with its corresponding value. A \texttt{<|ROW|>}  token is appended to mark the end of the flow.

Below is an example of single-processed rows before numerical encoding, illustrating the consistency of the format across different data points:

\texttt{<|SRCIP|>192.168.220.14<|SRCPORT|>49222<|DSTIP|>192.168.100.5\\<|DSTPORT|>443<|PROTOCOL|><TCP><|DURATION|>0.000<|BYTES|>0<|PKTS|>1\\<|FLAGS|><.AP...><|FLOWS|>1<|ROLE|><NORMAL><|CLASS|><NORMAL>\\<|DTIME|>0.0<|ROW|>
}

Finally, the entire text corpus is encoded into a single, continuous sequence of numerical token IDs. The encoder processes each line, mapping the structural and fixed categorical tokens to their unique IDs and applying the learned BPE model to encode the variable data values. This process results in a single \verb|NumPy|  array of \verb|int64| token IDs, which serves as the final input for training the foundation model. This entire process is highly efficient, capable of tokenising over 31 million flows in 4 hours and 44 minutes on our hardware, producing a final corpus of over 1 billion tokens.

\section{Results and Evaluations}
Following the application of our methodology to the full CIDDS-001 dataset, we evaluated the efficiency and effectiveness of the tokenisation process. The primary goal of this evaluation is to quantify the degree of data compression achieved and to assess the practical feasibility of the approach on a large-scale, real-world dataset. These results are crucial as they directly impact the computational resources required for training a subsequent foundation model. The decoding process was tested on the encoded data, achieving 100\% accuracy.

\subsection{Tokenisation Efficiency and Compression}
The entire preprocessed corpus, consisting of 31,287,933 lines (one for each NetFlow), was encoded into a numerical sequence. The key performance metrics of this process are summarised below:
\begin{table}[h]
    \centering
    \begin{tabular}{|l|r|}
        \hline
        \textbf{Description} & \textbf{Value} \\ \hline
        Total Flows Processed & 31,287,933 \\ \hline
        Total Characters (Input) & 6,284,718,381 \\ \hline
        Total Tokens (Output) & 1,016,754,652 \\ \hline
        Compression Ratio (Characters per Token) & 6.18 \\ \hline
    \end{tabular}
    \caption{Summary of Data Processing Metrics}
    \label{tab:data_processing_metrics}
\end{table}

The results demonstrate a significant data compression, with the tokenised sequence being substantially smaller than the raw character representation. The achieved compression ratio of approximately 6.18 indicates that, on average, more than six characters from the input text are represented by a single token. This is a vital outcome for training sequence-based models, as it significantly reduces the length of input sequences without information loss, resulting in lower memory consumption and a substantial reduction in the computational complexity of the attention mechanism in Transformer architectures.

Furthermore, the entire encoding process for over 31 million flows was completed in 4 hours and 44 minutes. This processing time underscores the practical viability of our method for handling datasets of a scale commonly encountered in network security and other real-world domains.

\section{Contribution and Next Steps}
The successful development and application of this hybrid tokeniser represents a significant achievement and our first primary contribution. This tokenisation approach is novel in its own right and, while demonstrated here on NetFlow data, is designed to be generalisable to other tabular data sources. By effectively embedding the structural and semantic information of tabular data into a compressed, sequential format, we have created a viable corpus for training a large-scale foundation model. 

The performance and efficiency demonstrated in this chapter are not merely preliminary results; they establish the feasibility of our approach. In the subsequent chapter, we will leverage this tokenised dataset to train our foundation model, exploring its capacity to learn meaningful representations of network traffic and its effectiveness in downstream security tasks.

\bibliographystyle{unsrtnat}
\bibliography{references}  






\end{document}